\documentclass[runningheads]{llncs}
\usepackage{graphicx}
\usepackage{amsmath}
\usepackage{tabularx}
\usepackage[dvipsnames]{xcolor}
\usepackage{hyperref}
\hypersetup{
    colorlinks=true,
    linkcolor=BurntOrange,
    citecolor=blue,
    filecolor=magenta,      
    urlcolor=blue,
}
\newcolumntype{M}[1]{>{\centering\arraybackslash}m{#1}}

\DeclareMathOperator*{\argmax}{argmax}
\begin{document}
\title{Large Scale Indexing of Generic Medical Image Data using Unbiased Shallow Keypoints and Deep CNN Features}
\titlerunning{\emph{Large Scale Indexing of Generic Medical Image Data}} 
%
\author{Laurent Chauvin\inst{1} \and Mohsen Ben Lazreg\inst{1} \and Jean-Baptiste Carluer\inst{2} \and William Wells III\inst{3,4} \and Matthew Toews\inst{1}}
\authorrunning{\emph{L. Chauvin et al.}}
%
\institute{École de Technologie Supérieure, Montreal, Canada \and
Université de Nantes, Nantes, France \and
Harvard Medical School, Boston, USA \and Massachusetts Institute of Technology, Boston, USA}
\maketitle              
\begin{abstract}
We propose a unified appearance model accounting for traditional shallow (i.e. 3D SIFT keypoints) and deep (i.e. CNN output layers) image feature representations, encoding respectively specific, localized neuroanatomical patterns and rich global information into a single indexing and classification framework. A novel Bayesian model combines shallow and deep features based on an assumption of conditional independence and validated by experiments indexing specific family members and general group categories in 3D MRI neuroimage data of 1010 subjects from the Human Connectome Project, including twins and non-twin siblings. A novel domain adaptation strategy is presented, transforming deep CNN vectors elements into binary class-informative descriptors. A GPU-based implementation of all processing is provided. State-of-the-art performance is achieved in large-scale neuroimage indexing, both in terms of computational complexity, accuracy in identifying family members and sex classification.
\end{abstract}
\section{Introduction}
Medical image repositories such as modern hospital picture archiving and communications systems (PACS) store increasingly large diverse 3D patient anatomy, e.g. diagnostic CT, MRI scans, histology, etc. In order to fully exploit these data via machine learning, it is important to model both specific instances, e.g. individuals and family members in personalised medicine, in addition to characteristics shared between broader group categories, e.g. males and females. What would be the most suitable computational approach, or data representation for such an approach?\\
\noindent
In the computer vision literature, deep convolutional networks (CNNs)~\cite{lecun1989backpropagation,krizhevsky2012imagenet} excel at classifying broad image categories, particularly where the number of training data samples $N$ is large relative to the number of classes labels of interest, e.g. the ImageNet dataset~\cite{deng2009imagenet} consisting of 1000 generic object categories with 1000 photographs each~\cite{Szegedy2016,He2016,Huang2017,Howard2017MobileNets:Applications,Chollet2017Xception:Convolutions,ZophLearningRecognition,ZophLearningRecognition,SzegedyInception-v4Learning}. Shallow keypoint based representations remain highly effective in the case of instance-retrieval where there may be at most a handful of examples per category~\cite{boiman2008defense,chandrasekhar2016practical,zheng2017sift,brendel2019approximating}. Additionally, keypoint correspondences can be used to achieve robust spatial alignment~\cite{Toews2013ab} required prior to classification via deep network, which are generally not invariant to rotation or scale changes. In the case of 3D brain MRI for example, keypoint-based methods have been used to identify individuals and family members with high accuracy from large, multi-modal datasets~\cite{kumar2018multi,Chauvin2020NeuroimageRelatives}.\\
\noindent
How do deep network activations and shallow keypoint descriptors compare in the context of medical image data indexing and classification? Can they be combined in a complementary, synergistic fashion? We address these questions in the context of a novel data model combining both shallow/early and deep/late CNN information, where each image is represented as 1) a variable-sized set of generic 3D SIFT keypoints and 2) a fixed-sized vector of deep CNN activations from networks pre-trained on generic visual object categories. In our model, shallow convolution maxima are used to establish robust spatial alignment, e.g. to a suitable atlas reference space. Additionally, we propose a novel information-theory based scheme to adapt CNN filters derived from generic objects to 3D medical image data, where CNN vectors are binarized according to element-wise threshold maximixing the mutual information between binary activation and class label. Our work builds about memory-based models~\cite{Toews2015,Chauvin2019AnalyzingManifold}, where image data are stored and used in on-the-fly kernel density estimation, an approach that approaches optimal Bayes error as the number of training data becomes large~\cite{Cover1967}, i.e. big data context.\\
\noindent
Experiments on 1010 MRIs from the Human Connectome Project~\cite{van2013wu}. Shallow keypoint descriptors found to be individually most informative for family member indexing and retrieval, although combined deep and shallow information leads to the highest overall performance, improving upon the keypoint signature method~\cite{Chauvin2020NeuroimageRelatives} that was used to discover previously unknown subject labelling errors in OASIS~\cite{marcus2007open}, ADNI~\cite{jack2008alzheimer} and HCP~\cite{van2013wu} public datasets widely used by the neuroimaging community. For the task of group classification, here male-female, both shallow and deep descriptors show similar performance individually and their combination results in a slight increase in AUC (area under the curve) performance.

\section{Related Work}

Our work seeks to combine generic shallow keypoint methods~\cite{Lowe2004,Toews2013ab} and CNN technnology~\cite{Szegedy2016,He2016,Huang2017,Howard2017MobileNets:Applications,Chollet2017Xception:Convolutions,ZophLearningRecognition,ZophLearningRecognition,SzegedyInception-v4Learning} for the purpose of large-scale generic indexing and classification of medical image data.\\
\noindent
A variety of context- and task-specific approaches may be used to detect and describe keypoints~\cite{yi2016lift,ono2018lf}, and deep filter responses excel for few-label-many-data contexts such as group classification~\cite{zheng2017sift}. Nevertheless, comparisons have shown variants parametric (or 'hand-crafted') descriptors such as gradient orientation histograms can be more effective, particularly for matching specific visual scenes~\cite{schonberger2017comparative,dong2015domain} or retrieving specific object instance retrieval~\cite{zheng2017sift}. A possible explanation is that bias is introduced in early (shallow) filtering layers~\cite{geirhos2018imagenet,ringer2019texture}. For example, input filters derived from stochastic backpropagation are generally non-symmetric and biased to specific image oriented image patterns~\cite{lecun1989backpropagation,krizhevsky2012imagenet}, unlike rotationally symmetric operators such as the difference-of-Gaussian or uniformly sampled Gaussian derivative filters used in SIFT descriptors which are not biased to any particular orientation~\cite{Lowe2004,Toews2013ab}.\\
\noindent
The importance of shallow filter information has been recently demonstrated~\cite{brendel2019approximating}. 3D keypoint indexing demonstrated state-of-the-art performance in identifying individuals and family members from large sets of brain MRI data~\cite{Chauvin2020NeuroimageRelatives}, but also identifying previously unknown subject labeling errors in widely neuroimaging datasets. Keypoint and CNN representations can be combined resulting in superior performance for instance identification~\cite{chandrasekhar2016practical}.

\section{Method}

\begin{figure}[!ht]
  \centering
   \includegraphics[width=1\textwidth]{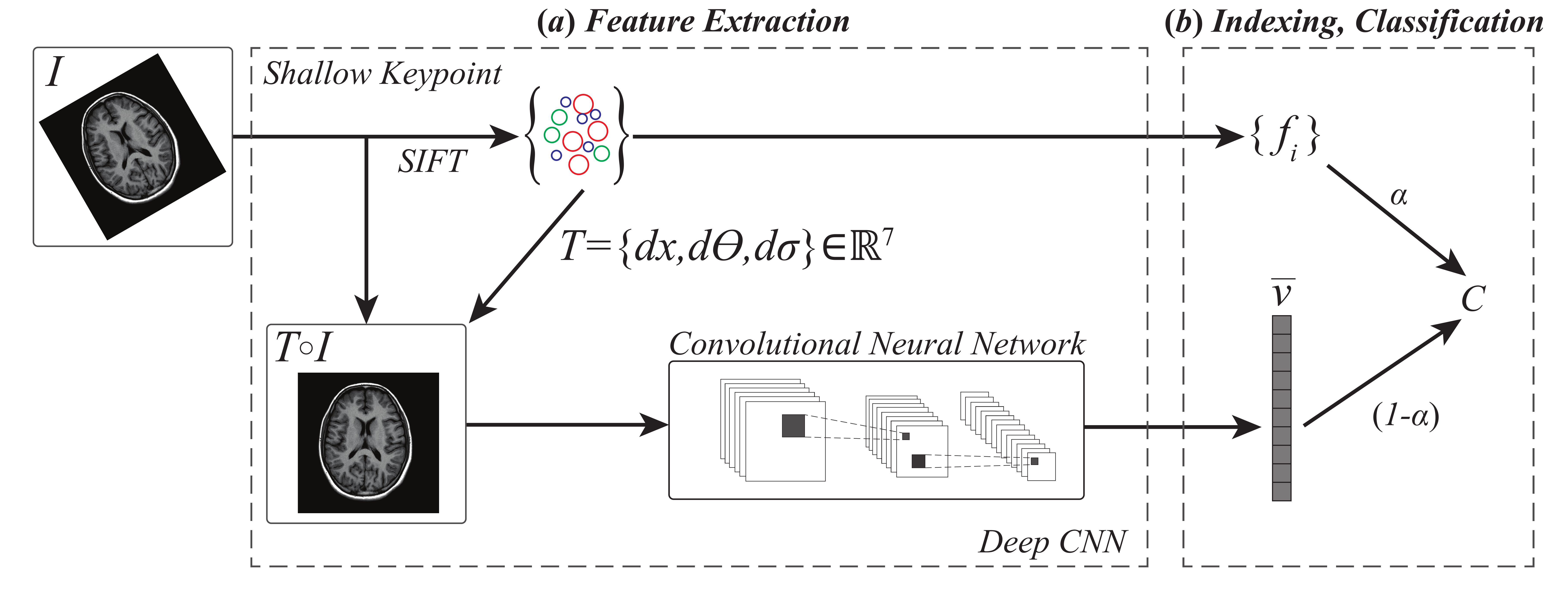}
   \caption{Illustrating the extraction of shallow (early) keypoint information and deep CNN. Keypoints $\{f_j\}$ are extracted from a generally unaligned image $I$. A robust global similarity transform $T$ between the image and a reference atlas determined via the keypoint-based alignment method~\cite{Toews2013a,toews2013feature}. Spatial alignment is applied to the image $T\circ I$ after which a deep CNN response vector $\bar{v}$ is sampled from layers near the output of a pre-trained CNN. Resampling the image according to a global transform prior to CNN feature extraction is similar to a spatial transformer network~\cite{JaderbergSpatialNetworks}.}
  \label{fig:keypoint_cnn}
\end{figure}

\begin{figure}[!h]
  \centering
  \includegraphics[width=0.5\textwidth]{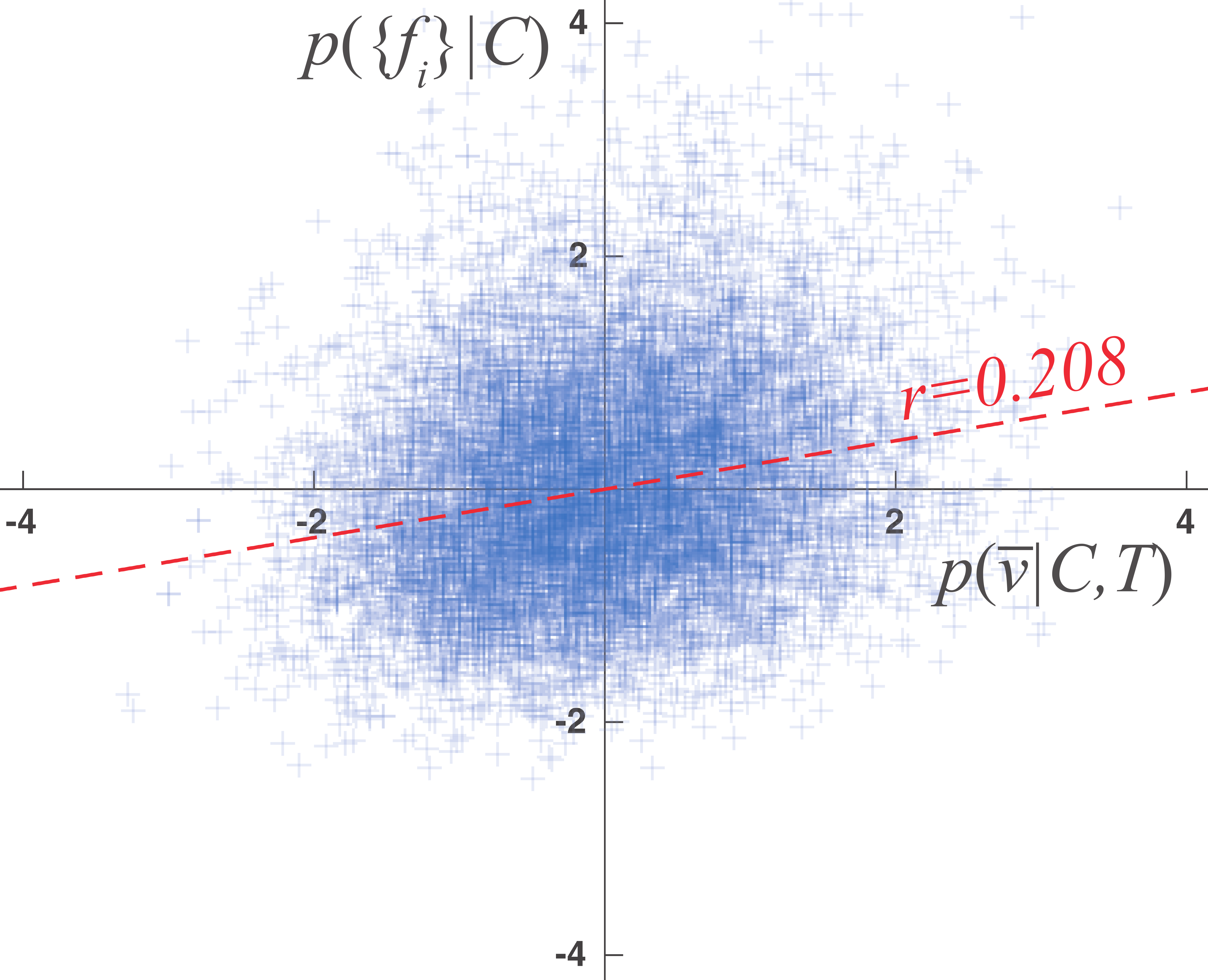}
  \caption{A plot of the distribution of shallow $p(\{f_{i}\}|C,T)$ and deep $p(\bar{v}|C,T)$ data pairs shows a very low correlation coefficient ($r=0.208$), indicating high degree of independence between these two information sources.}
  \label{fig:scatter_sift_cnn}
\end{figure}

We seek a model combining shallow, variable-sized local keypoint sets $\{f_i\}$ and fixed-length deep descriptor $\bar{v}$ into a generic system suitable for both indexing and classification. To this end, we consider an instance or memory-based model based on adaptive kernel density estimation, combining generic shallow and deep CNN information. Figure~\ref{fig:keypoint_cnn} illustrates our approach, where (a) represents generic, existing feature extraction technology and (b) represents the model we investigate here.
\noindent
We propose a Bayesian a maximum a posteriori (MAP) formulation, where the goal is maximize the posterior probability over a discrete random variable of class $C=\{C_k\}$ conditioned on data $(\{f_i\},\bar{v})$ and spatial transform $T$ defined as:
\begin{align}
    p(C|\{f_i\},\bar{v},T) &\propto p(\{f_i\},\bar{v}|C,T)p(C|T) \label{eq:bayes}\\
    &= p(\{f_i\}|C,T)p(\bar{v}|C,T)p(C|T) \label{eq:cond_ind}\\
    &\approx p(\{f_i\}|C)p(\bar{v}|C,T) \label{eq:ind_trans}
\end{align}
In Equation~\eqref{eq:bayes} the proportionality results from Bayes rule, where $p(\{f_i\},\bar{v}|C,T)$ is the probability of feature data $(\{f_i\},\bar{v})$ conditioned on class $C$ and image-to-atlas transform $T$, and $p(C|T)$ is a prior probability of class $C$ conditioned on transform which we assume to be uniform (uninformative) in the current analysis. The equality~\eqref{eq:cond_ind} is based on the assumption of conditional independence between shallow keypoint $\{f_i\}$ and deep CNN $\bar{v}$ information given class and transform $(C,T)$. Furthermore, keypoint descriptors $\{f_i\}$ are conditionally independent on 3D similarity transform $T$ given class $C$, due to geometrical invariance i.e. $p(\{f_i\}|C,T)=p(\{f_i\}|C)$ in equation~\eqref{eq:ind_trans}, while CNN descriptors are not invariant to scale and rotation changes and require spatial normalization.\\
\noindent
The fusion of deep and shallow features was based on equation~\eqref{eq:bayes} and~\eqref{eq:ind_trans}, by taking the log of $p(C|\{f_{i}\},\bar{v},T)$ and weighting each term with a fusion parameter $\alpha$ such as
\begin{align}
    \log p(C|\{f_{i}\},\bar{v},T) \propto \alpha \log p(\{f_{i}\}|C) + (1-\alpha) \log p(\bar{v}|C,T)
    \label{eq:fusion}
\end{align}
with the first term being proportional to the log Jaccard keypoint signature distance proposed in~\cite{Chauvin2020NeuroimageRelatives}. In Equation~\eqref{eq:fusion}, parameter $\alpha=[0,1]$ is an exponent factor balancing the relative variances of conditionally independent shallow and deep feature distributions, empirically determined. The assumption of conditional independence is validated from data in Figure~\ref{fig:scatter_sift_cnn}, where shallow and deep features are virtually uncorrelated, to our knowledge the first time this result is reported. \\
\noindent
The specific formulation of each term in equation~\eqref{eq:fusion} is as follows:
$p(\{f_i\}|C)$ represents the probability of keypoint set $\{f_i\}$ conditional on $C$ is modeled as the product of independent and identically distributed keypoints
\begin{align}
    p(\{f_i\}|C) = \prod_i p(f_i|C),
    \label{eq:cond_indep}
\end{align}
where the density associated with an individual keypoint $p(f_i|C)$ is expressed as proportional to the marginalisation over a training keypoint set $\{f_j\}$:
\begin{align}
    p(f_i|C_k) &\propto \frac{1}{N_{k,f}}\sum_j p(f_i|f_j,C_k)p(f_j|C_k) + 1, \\
    &= \frac{1}{N_{k,f}} \sum_j \exp\left(-\frac{\|f_i-f_j\|^2}{1+dNN_i^2}\right)[C_j=C_k] + 1
\end{align}
where $N_{k,f}=|\{f_j: C_{j} = C_{k}\}|$ is the number of training keypoints $f_j$ with class label $C_k$,
$p(f_i|f_j,C_k)=\exp\left(-\frac{\|f_i-f_j\|^2}{1+dNN_i^2}\right)$ is an exponential kernel with variance defined by $dNN_i = \min_{j}{\|f_i-f_j\|}$ and $p(f_j|C_k)=[C_j=C_k]$ is the Iverson bracket evaluating to 1 if label $C_j$ associated with $f_j$ is equivalent to label $C_k$ and zero otherwise. The $+ 1$ term is included to represent the contribution of uniform background noise, and ensures the product in Equation~\eqref{eq:cond_indep} does not vanish.\\
\noindent
In a similar fashion, the term $p(\bar{v}|C_k,T)$ in equation~\eqref{eq:fusion} is defined as 
\begin{align}
p(\bar{v}|C_k,T) &\propto \frac{1}{N_{k,v}} \sum_{j}  p(\bar{v}|\bar{v}_{j},C_k)p(\bar{v}_{j}|C_{k}) + 1\\
             &\propto \frac{1}{N_{k,v}} \sum_{j} \exp \left(-\frac{||\bar{v}-\bar{v}_{j}||^2}{\mu^{2}}\right)[C_{j}=C_{k}] + 1
\end{align}
where $N_{k,v}=|\{\bar{v}_{j}: C_{j} = C_{k}\}|$ is the number of training vectors $\bar{v}_j$ with class label $C_k$, $p(\bar{v}|\bar{v}_{j},C)=\exp \left(-\frac{||\bar{v}-\bar{v}_{j}||^2}{\mu^{2}}\right)$ with $\mu=\frac{1}{N_{k,v}-1}\sum_j\|\bar{v}-\bar{v}_{j}\|$ the mean distance between feature vector $\bar{v}$ each training feature vector $\bar{v}_{j}$, and $p(\bar{v}_{j}|C_{k})= [C_{j} = C_{k}]$ is the Iverson bracket.

\subsection{CNN Vectors and Domain Adaptation}
In order to develop a generally applicable CNN implementation, we used the strategy of transfer learning, adapting deep CNN information in the form of generic models pre-trained on ImageNet dataset~\cite{deng2009imagenet}, consisting of 1,000,000 images of 1000 object classes. While specialized training on domain-specific data may improve results, large pre-trained CNNs are commonly used as generic feature extractors and are surprisingly effective in medical image analysis where data may be scarce~\cite{tajbakhsh2016convolutional}.\\ 
\noindent
We adopt a novel domain adaptation scheme based on information theory, where a raw vector of CNN information $\bar{v}$ extracted from 2D ROI is converted into an informative, lightweight binary vector. Each element $\bar{v}[i]$ is binarized by a threshold $\tau_i$, where $\tau_i$ is determined such that the mutual information (or information gain) between binary element $\bar{v}[i]$ and the class of interest $C$ is maximized, as in classic decision tree training~\cite{breiman1984classification}:
\begin{align}
    \tau_i = \argmax_{\tau} \left\{ H_{\tau}(\bar{v}[i]) - H_{\tau}(\bar{v}[i]|C) \right\},
\end{align}
where $H_{\tau}(\bar{v}[i])$ and $H_{\tau}(\bar{v}[i]|C)$ are the binary entropy and binary conditional entropy (Shannon), estimated from training samples. Note that in the case of sparse data, e.g. 1 sample per class, the conditional entropy $H_{\tau}(\bar{v}[i]|C)=0$ and $\tau_i$ maximizes the entropy. 

\section{Experiments}
\begin{figure}[!ht]
  \centering
   \includegraphics[width=1\textwidth]{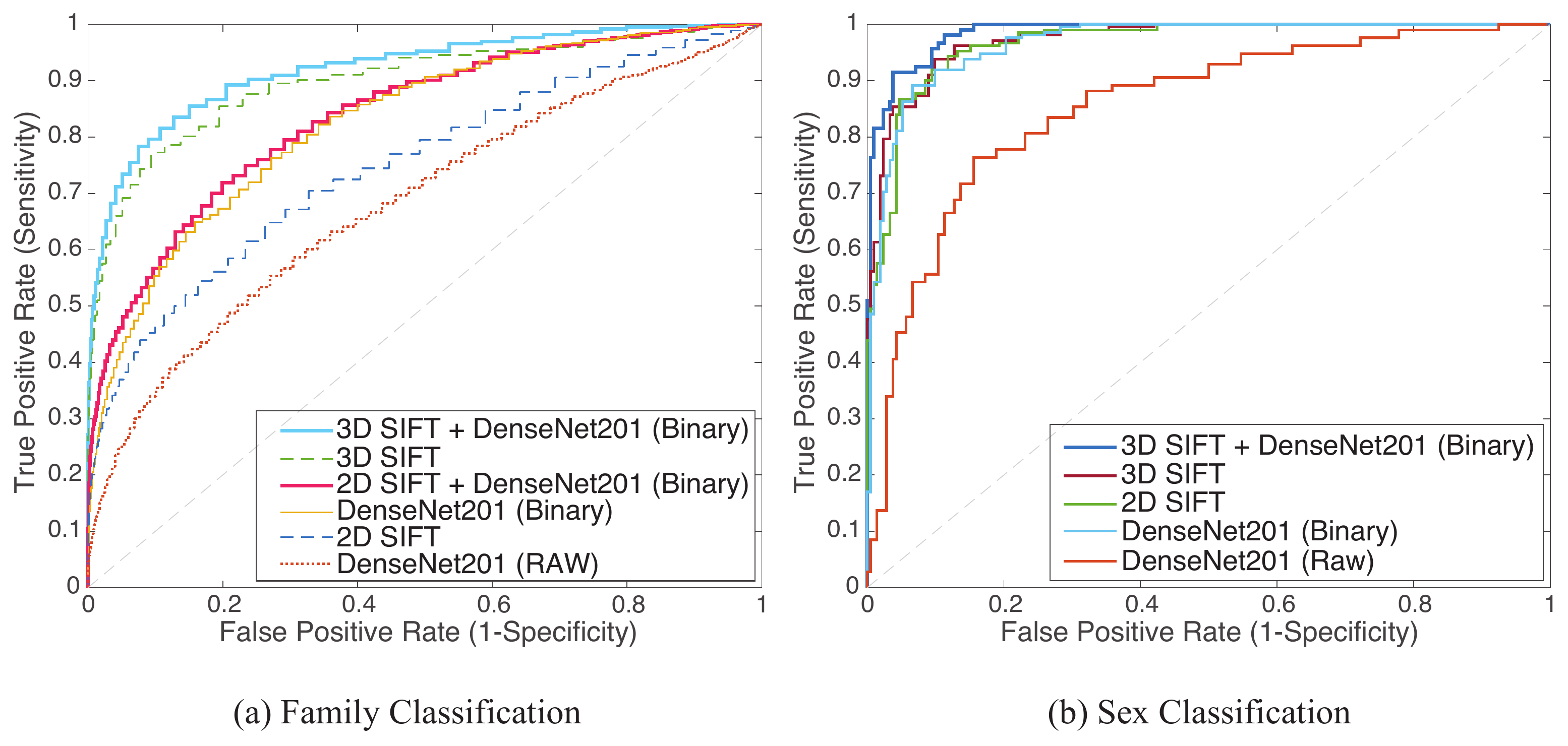}
   \caption{Receiver Operating Characteristic (ROC) curves for (a) family and (b) sex classification with DenseNet201, 2D SIFT, 3D SIFT and combination of features. Area Under the Curve (AUC) are reported in Table~\ref{tab:Family_AUC}.}
  \label{fig:mixing_results_family_sex_class}
\end{figure}

The goal of experiments was to evaluate the effectiveness of shallow and deep feature combinations in two classification contexts: 1) many-class-few-data (e.g. few-shot learning) and 2) few-class-many-data (e.g. group classification). Shallow keypoints $\{f_i\}$ are extracted with the 2D and 3D SIFT-Rank method based on a GPU implementation, with a memory footprint 150x smaller than the original image, deep feature vectors $\bar{v}$ are taken from DenseNet201~\cite{Huang2017}. CNN and 2D SIFT features are sampled from a single 2D mid-axial MRI in each 3D volume, additional slices or volumetric CNN information could potentially improve classification, however a single slice evaluation protocol is similar to diagnostic MRI data which often consist of a small number of high-res mid-axial slices. 10 different CNN models have been evaluated (InceptionV3~\cite{Szegedy2016}, ResNet50~\cite{He2016}, DenseNet121,169,201~\cite{Huang2017}, MobileNet~\cite{Howard2017MobileNets:Applications}, Xception~\cite{Chollet2017Xception:Convolutions}, NASNetMobile~\cite{ZophLearningRecognition}, NASNetLarge~\cite{ZophLearningRecognition}, InceptionResNetV2~\cite{SzegedyInception-v4Learning}), but for clarity, only the best layer of the best model (here, layer 704 of binarized DenseNet201), identified through a semi-exhaustive evaluation has been preserved in the following experiments, resulting in a 1920-dimension feature vector per image.\\ \\
\noindent
\textbf{Specific Indexing (Family Member Labels)}:
Here, our dataset was composed of 1010 T1w 0.7mm isotropic images from the Human Connectome Project Q4 release, with 1 image per subject and regrouped in 439 families (including monozygotic, dizygotic twins and full-siblings). Subjects are aged between 22-36 years old (mean: 29), with 468 males and 542 females. A total of 1,488,065 SIFT keypoints have been extracted in 29min (approximately 2 secs/image) for a size of 90MB, and 1010 deep features. A total of $N(N-1)/2=1,020,100$ pairwise similarity scores based on equation~\eqref{eq:fusion} are computed via KD-tree indexing (FLANN~\cite{Muja2014}) in 0.35 sec/image on an Intel Xeon Silver 4110@2.10Ghz. Results presented in Figure~\ref{fig:mixing_results_family_sex_class} and Table~\ref{tab:Family_AUC} show that domain adaptation through binarization significantly improves performance and for few-shot learning classification, shallow keypoints perform slightly better than deep features, but best performance is achieved via a fusion of both, suggesting complementary information.\\ \\
\noindent
\textbf{General Group Classification (Sex Labels) }:
In this experiment, we used a subset of 424 images of the HCP dataset, to preserve only one family member per family, with an equal number of males and females in order to reduce the possible biases due to genetic influences or imbalanced groups. A total of 624,643 keypoints have been extracted in 12min for a total size of 38MB, and 424 deep features.
Results in Figure~\ref{fig:mixing_results_family_sex_class} and Table~\ref{tab:Family_AUC} still indicate a significant improvement through domain adaptation, but relatively similar performances with deep or shallow features. It should be noted that combining features also lead to an improvement in performances. 

\begin{table}[!h]
	\begin{center}
    \begin{tabular}{M{0.4\textwidth}M{0.2\textwidth}M{0.2\textwidth}}
        \hline
        \textbf{Model} & \textbf{Family Classification (AUC)} & \textbf{Sex Classification (AUC)}\\
        \hline
        \hline
        \textbf{3D SIFT + DenseNet201 (B)} & \textbf{0.9258} & \textbf{0.9875}\\
        3D SIFT & 0.9058 & 0.9712\\
        2D SIFT + DenseNet201 (B) & 0.8420 & -\\
        DenseNet201 (B) & 0.8305 & 0.9649\\
        2D SIFT & 0.7590 & 0.9652\\
        DenseNet201 & 0.6882 & 0.8586\\

        \hline
    \end{tabular}
    \caption{Area Under the Curve (AUC) for different model and combination for family and sex classification, based on Receiver Operating Characteristic (ROC) in Figure~\ref{fig:mixing_results_family_sex_class}.}
  	\label{tab:Family_AUC}
  \end{center}
\end{table}

\section{Discussion}
We propose a novel Bayesian formulation in order to combine generic keypoint and CNN information into a single, highly efficient memory-based model for indexing and classifying generic 3D medical image data. Our model is invariant to 3D similarity transforms and the keypoint extraction process is highly efficient and greatly reduce the memory footprint necessary to store images in memory (by a factor of 150), which proved to be very useful for large scale image analysis and classification. The approach presented here for domain adaptation of deep features lead to an increase in family and sex classification performances, particularly in the case of many-class-few-data indexing scenario of family member prediction.\\
\noindent
As our current work is based entirely on generic features requiring no specialised training, and the entire system can be applied as is to large medical image datasets. To date, we have processed 25K Brain MRIs from the UK Biobank~\cite{Sudlow2015} and 20K Lung CT COPD-Gene~\cite{regan2011genetic}, these datasets can now fit into standard RAM on a laptop computer, and an individual image query against data is well under 1 second. Kernel variance parameters can be estimated on-the-fly, CNN vector domain adaptation and further learning-based methods will be investigated against baseline generic performance, model of conditional independence appears to be a promising strategy supported by evidence here.

\section{Code References}
\begin{itemize}
    \setlength\itemsep{0.7em}
    \setlength{\itemindent}{.1in}
    \item[\textbullet] \href{http://www.matthewtoews.com/fba/featExtract1.6.tar.gz}{3D SIFT Binary Utilities}
    \item[\textbullet] \href{https://github.com/CarluerJB/3D_SIFT_PYCUDA}{GPU Implementation}
    \item[\textbullet] \href{https://github.com/3dsift-rank/3DSIFT-Rank}{Large-scale Matching}
    \item[\textbullet] \href{https://github.com/pEtienn/3D-SIFT-keypoints-utilities}{Python Utilities}
\end{itemize}

%
%
%
\bibliographystyle{splncs04}
\bibliography{MICCAI20,extra}

\begin{thebibliography}{10}
\providecommand{\url}[1]{\texttt{#1}}
\providecommand{\urlprefix}{URL }
\providecommand{\doi}[1]{https://doi.org/#1}

\bibitem{boiman2008defense}
Boiman, O., Shechtman, E., Irani, M.: In defense of nearest-neighbor based
  image classification. In: 2008 IEEE Conference on Computer Vision and Pattern
  Recognition. pp.~1--8. IEEE (2008)

\bibitem{breiman1984classification}
Breiman, L., Friedman, J., Olshen, R., Stone, C.: Classification and regression
  trees  (1984)

\bibitem{brendel2019approximating}
Brendel, W., Bethge, M.: Approximating cnns with bag-of-local-features models
  works surprisingly well on imagenet. arXiv preprint arXiv:1904.00760  (2019)

\bibitem{chandrasekhar2016practical}
Chandrasekhar, V., Lin, J., Morere, O., Goh, H., Veillard, A.: A practical
  guide to cnns and fisher vectors for image instance retrieval. Signal
  Processing  \textbf{128},  426--439 (2016)

\bibitem{Chauvin2019AnalyzingManifold}
Chauvin, L., Kumar, K., Desrosiers, C., De~Guise, J., Wells, W., Toews, M.:
  {Analyzing Brain Morphology on the Bag-of-Features Manifold}. In: Lecture
  Notes in Computer Science (including subseries Lecture Notes in Artificial
  Intelligence and Lecture Notes in Bioinformatics). vol. 11492 LNCS, pp.
  45--56. Springer Verlag (2019). \doi{10.1007/978-3-030-20351-1{\_}4}

\bibitem{Chauvin2020NeuroimageRelatives}
Chauvin, L., Kumar, K., Wachinger, C., Vangel, M., de~Guise, J., Desrosiers,
  C., Wells, W., Toews, M.: {Neuroimage signature from salient keypoints is
  highly specific to individuals and shared by close relatives}. NeuroImage
  \textbf{204}(20) (9 2020). \doi{10.1016/j.neuroimage.2019.116208}

\bibitem{Chollet2017Xception:Convolutions}
Chollet, F.: {Xception: Deep learning with depthwise separable convolutions}.
  In: Proceedings - 30th IEEE Conference on Computer Vision and Pattern
  Recognition, CVPR 2017. vol. 2017-January, pp. 1800--1807. Institute of
  Electrical and Electronics Engineers Inc. (11 2017).
  \doi{10.1109/CVPR.2017.195}

\bibitem{Cover1967}
Cover, T., Hart, P.: {Nearest neighbor pattern classification}. IEEE
  Transactions on Information Theory  \textbf{13}(1),  21--27 (1 1967).
  \doi{10.1109/TIT.1967.1053964},
  \url{http://ieeexplore.ieee.org/document/1053964/}

\bibitem{deng2009imagenet}
Deng, J., Dong, W., Socher, R., Li, L.J., Li, K., Fei-Fei, L.: Imagenet: A
  large-scale hierarchical image database. In: 2009 IEEE conference on computer
  vision and pattern recognition. pp. 248--255. Ieee (2009)

\bibitem{dong2015domain}
Dong, J., Soatto, S.: Domain-size pooling in local descriptors: Dsp-sift. In:
  Proceedings of the IEEE conference on computer vision and pattern
  recognition. pp. 5097--5106 (2015)

\bibitem{geirhos2018imagenet}
Geirhos, R., Rubisch, P., Michaelis, C., Bethge, M., Wichmann, F.A., Brendel,
  W.: Imagenet-trained cnns are biased towards texture; increasing shape bias
  improves accuracy and robustness. arXiv preprint arXiv:1811.12231  (2018)

\bibitem{He2016}
He, K., Zhang, X., Ren, S., Sun, J.: {Deep residual learning for image
  recognition}. In: Proceedings of the IEEE Computer Society Conference on
  Computer Vision and Pattern Recognition (2016). \doi{10.1109/CVPR.2016.90}

\bibitem{Howard2017MobileNets:Applications}
Howard, A.G., Zhu, M., Chen, B., Kalenichenko, D., Wang, W., Weyand, T.,
  Andreetto, M., Adam, H.: {MobileNets: Efficient Convolutional Neural Networks
  for Mobile Vision Applications}  (4 2017),
  \url{http://arxiv.org/abs/1704.04861}

\bibitem{Huang2017}
Huang, G., Liu, Z., Van Der~Maaten, L., Weinberger, K.Q.: {Densely connected
  convolutional networks}. In: Proceedings - 30th IEEE Conference on Computer
  Vision and Pattern Recognition, CVPR 2017 (2017). \doi{10.1109/CVPR.2017.243}

\bibitem{jack2008alzheimer}
Jack~Jr, C.R., Bernstein, M.A., Fox, N.C., Thompson, P., Alexander, G., Harvey,
  D., Borowski, B., Britson, P.J., L.~Whitwell, J., Ward, C., {others}: {The
  Alzheimer's disease neuroimaging initiative (ADNI): MRI methods}. Journal of
  Magnetic Resonance Imaging: An Official Journal of the International Society
  for Magnetic Resonance in Medicine  \textbf{27}(4),  685--691 (2008)

\bibitem{JaderbergSpatialNetworks}
Jaderberg, M., Simonyan, K., Zisserman, A., Kavukcuoglu, K.: {Spatial
  Transformer Networks}. Tech. rep.,
  \url{http://papers.nips.cc/paper/5854-spatial-transformer-networks}

\bibitem{krizhevsky2012imagenet}
Krizhevsky, A., Sutskever, I., Hinton, G.E.: Imagenet classification with deep
  convolutional neural networks. In: Advances in neural information processing
  systems. pp. 1097--1105 (2012)

\bibitem{kumar2018multi}
Kumar, K., Toews, M., Chauvin, L., Colliot, O., Desrosiers, C.: Multi-modal
  brain fingerprinting: a manifold approximation based framework. NeuroImage
  \textbf{183},  212--226 (2018)

\bibitem{lecun1989backpropagation}
LeCun, Y., Boser, B., Denker, J.S., Henderson, D., Howard, R.E., Hubbard, W.,
  Jackel, L.D.: Backpropagation applied to handwritten zip code recognition.
  Neural computation  \textbf{1}(4),  541--551 (1989)

\bibitem{Lowe2004}
Lowe, D.G.: {Distinctive image features from scale-invariant keypoints}.
  International Journal of Computer Vision  \textbf{60}(2),  91--110 (2004).
  \doi{10.1023/B:VISI.0000029664.99615.94}

\bibitem{marcus2007open}
Marcus, D.S., Wang, T.H., Parker, J., Csernansky, J.G., Morris, J.C., Buckner,
  R.L.: {Open Access Series of Imaging Studies (OASIS): cross-sectional MRI
  data in young, middle aged, nondemented, and demented older adults}. Journal
  of cognitive neuroscience  \textbf{19}(9),  1498--1507 (2007)

\bibitem{Muja2014}
Muja, M., And, D.L.I.t.o.p.a., 2014, U.: {Scalable nearest neighbor algorithms
  for high dimensional data}. ieeexplore.ieee.org  (2014),
  \url{http://ieeexplore.ieee.org/abstract/document/6809191/}

\bibitem{ono2018lf}
Ono, Y., Trulls, E., Fua, P., Yi, K.M.: Lf-net: learning local features from
  images. In: Advances in Neural Information Processing Systems. pp. 6234--6244
  (2018)

\bibitem{regan2011genetic}
Regan, E.A., Hokanson, J.E., Murphy, J.R., Make, B., Lynch, D.A., Beaty, T.H.,
  Curran-Everett, D., Silverman, E.K., Crapo, J.D.: Genetic epidemiology of
  copd (copdgene) study design. COPD: Journal of Chronic Obstructive Pulmonary
  Disease  \textbf{7}(1),  32--43 (2011)

\bibitem{ringer2019texture}
Ringer, S., Williams, W., Ash, T., Francis, R., MacLeod, D.: Texture bias of
  cnns limits few-shot classification performance. arXiv preprint
  arXiv:1910.08519  (2019)

\bibitem{schonberger2017comparative}
Schonberger, J.L., Hardmeier, H., Sattler, T., Pollefeys, M.: Comparative
  evaluation of hand-crafted and learned local features. In: Proceedings of the
  IEEE Conference on Computer Vision and Pattern Recognition. pp. 1482--1491
  (2017)

\bibitem{Sudlow2015}
Sudlow, C., Gallacher, J., Allen, N., Beral, V., Burton, P., Danesh, J.,
  Downey, P., Elliott, P., Green, J., Landray, M., Liu, B., Matthews, P., Ong,
  G., Pell, J., Silman, A., Young, A., Sprosen, T., Peakman, T., Collins, R.:
  {UK Biobank: An Open Access Resource for Identifying the Causes of a Wide
  Range of Complex Diseases of Middle and Old Age}. PLOS Medicine
  \textbf{12}(3),  e1001779 (3 2015). \doi{10.1371/journal.pmed.1001779},
  \url{http://dx.plos.org/10.1371/journal.pmed.1001779}

\bibitem{SzegedyInception-v4Learning}
Szegedy, C., Ioffe, S., Vanhoucke, V., on, A.A.T.f.A.c., 2017, u.:
  {Inception-v4, inception-resnet and the impact of residual connections on
  learning}. aaai.org
  \url{https://www.aaai.org/ocs/index.php/AAAI/AAAI17/paper/viewPaper/14806}

\bibitem{Szegedy2016}
Szegedy, C., Vanhoucke, V., Ioffe, S., Shlens, J., Wojna, Z.: {Rethinking the
  Inception Architecture for Computer Vision}. In: Proceedings of the IEEE
  Computer Society Conference on Computer Vision and Pattern Recognition
  (2016). \doi{10.1109/CVPR.2016.308}

\bibitem{tajbakhsh2016convolutional}
Tajbakhsh, N., Shin, J.Y., Gurudu, S.R., Hurst, R.T., Kendall, C.B., Gotway,
  M.B., Liang, J.: Convolutional neural networks for medical image analysis:
  Full training or fine tuning? IEEE transactions on medical imaging
  \textbf{35}(5),  1299--1312 (2016)

\bibitem{Toews2015}
Toews, M., Wachinger, C., Estepar, R.S.J., Wells, W.M.r.: {A Feature-Based
  Approach to Big Data Analysis of Medical Images.} Information processing in
  medical imaging : proceedings of the ... conference  \textbf{24},  339--350
  (2015)

\bibitem{Toews2013ab}
Toews, M., Wells, W.M.: {Efficient and robust model-to-image alignment using 3D
  scale-invariant features}. Medical Image Analysis  \textbf{17}(3),  271--282
  (2013). \doi{10.1016/j.media.2012.11.002}

\bibitem{Toews2013a}
Toews, M., Wells, W.M.: {Efficient and robust model-to-image alignment using 3D
  scale-invariant features}. Medical Image Analysis  \textbf{17}(3),  271--282
  (2013). \doi{10.1016/j.media.2012.11.002}

\bibitem{toews2013feature}
Toews, M., Z{\"{o}}llei, L., Wells, W.M.: {Feature-based alignment of
  volumetric multi-modal images}. In: International Conference on Information
  Processing in Medical Imaging. pp. 25--36. Springer (2013)

\bibitem{van2013wu}
Van~Essen, D.C., Smith, S.M., Barch, D.M., Behrens, T.E.J., Yacoub, E.,
  Ugurbil, K., Consortium, W.M.H.C.P., {others}: {The WU-Minn human connectome
  project: an overview}. Neuroimage  \textbf{80},  62--79 (2013)

\bibitem{yi2016lift}
Yi, K.M., Trulls, E., Lepetit, V., Fua, P.: Lift: Learned invariant feature
  transform. In: European Conference on Computer Vision. pp. 467--483. Springer
  (2016)

\bibitem{zheng2017sift}
Zheng, L., Yang, Y., Tian, Q.: Sift meets cnn: A decade survey of instance
  retrieval. IEEE transactions on pattern analysis and machine intelligence
  \textbf{40}(5),  1224--1244 (2017)

\bibitem{ZophLearningRecognition}
Zoph, B., Brain, G., Vasudevan, V., Shlens, J., Le~Google~Brain, Q.V.:
  {Learning Transferable Architectures for Scalable Image Recognition}. Tech.
  rep.

\end{thebibliography}

\end{document}